\documentclass[11pt,a4paper]{article}
\usepackage{times,latexsym}
\usepackage{url}
\usepackage[T1]{fontenc}

\usepackage[usenames,dvipsnames,svgnames,table]{xcolor}
\definecolor{purple}{RGB}{204, 0, 255}
\definecolor{green}{RGB}{0, 230, 0}

\usepackage{graphicx}
\usepackage{subcaption}
\usepackage{amsmath}
\usepackage{amsfonts}
\usepackage{color}

\usepackage[]{tacl2018v2}

\taclpubformattrue
\usepackage{xspace,mfirstuc,tabulary}

\newif\iftaclinstructions
\taclinstructionsfalse 
\iftaclinstructions

\renewcommand{\confidential}{}
\renewcommand{\anonsubtext}{(No author info supplied here, for consistency with
TACL-submission anonymization requirements)}
\newcommand{\instr}
\fi

\iftaclpubformat 

\else

\fi

\title{A Self-Explainable Stylish Image Captioning  Framework via Multi-References}

\author{
 Chengxi Li\\
 University of Kentucky\\
 {\sf cli289@uky.edu}\\

 Brent Harrison\\
 University of Kentucky\\
 {\sf harrison@cs.uky.edu}\\
}
\date{}

\begin{document}
\maketitle
\begin{abstract}
In this paper, we propose to build a stylish image captioning model through a Multi-style Multi modality mechanism (2M). We demonstrate that with 2M, we can build an effective stylish captioner and that multi-references produced by the model can also support explaining the model through
identifying erroneous input features on faulty examples. We show how this 2M mechanism can be used to build stylish captioning models and show how these models can be utilized to provide explanations of likely errors in the models. 
\end{abstract}

\section{Introduction}

While classic image captioning approaches show deep understanding of image composition and language construction, they often lack elements that make communication distinctly human. 
To address this issue, some researchers have tried to add personality to image captioning in order to generate \emph{stylish} captions.
In general, stylish captioning systems are divided into two categories based on how they are trained: 
single style and multi-style. 
Single-style training involves training one model for each personality, whereas multi-style techniques learn to generate captions in many different styles using one model. 


Past attempts at generating multi-style captioners, such as~\cite{shuster2019engaging}, have struggled, likely because they require greater knowledge about the input image as compared to single style captioners. 
One way to address this is to utilize multi-modality~\cite{zhang2021vinvl}; however, many times 
the multi-modal
approaches focus on generating features that describe local visual inputs, rather than a global context. 

We attempt to address this limitation of multi-modality features by allowing the model to self-select the most salient features from both visual and textual features according to the current global context. 
Specifically, we use ResNext features, as used by Shuster \textit{et al.}'s work describing global features, as well as region-based dense caption features generated by the DenseCap network~\cite{densecap} for local features.
into natural language. 

One issue with this approach is that both ResNext and dense caption features are generated using pre-trained networks.
This increases the likelihood that our generated text is erroneous, since these models were not trained for our specific problem. 
In situations like this, it is beneficial for the model to be able to \textit{explain} the source of said erroneous text generation so that a human operator can work to correct them.  Through experiments, we discover that our proposed multi-style model with multi-modality (image+text) inputs, which we refer to as \textit{\textbf{2M}},  could
be used to produce \textit{\textbf{multi-references}} (in text), which are easily interpreted by humans and can help identify the source of errors, such as poor style training or if the errors result from ResNext, or dense captions, present in the stylish captions. 
 
With all these considerations, we show the value of our approach by building two stylish captioners: 
one using a Multi-UPDOWN captioner, and another created by fine tuning a multimodal transformer. We also use the multi-references produced by these models to construct a multi-view tree which can be used to generate explanations for any errors present in generated text.  We design our evaluations to answer two questions: 1. Can the 2M concept help to build an effective multi-style stylish caption model? 2. Can the 2M concept help to explain errors in the model?  We evaluate stylish caption model's performance using various quantitative NLP metrics, 
and we perform
a qualitative analysis 
to evaluate
the overall expressiveness and diversity of generated captions. 
Secondly, we evaluate how 2M could help us identify feature errors. We perform the quantitative  evaluation and examine the predicting accuracy on two stylish capioners where we built with multi-UPDOWN models and transformer. We also perform  qualitative  evaluation on multiple datasets by walking through examples and demonstrate the multi-references are helpful 
in explaining the model to humans.
\begin{figure*}[t]
\center
\includegraphics[scale=0.49]{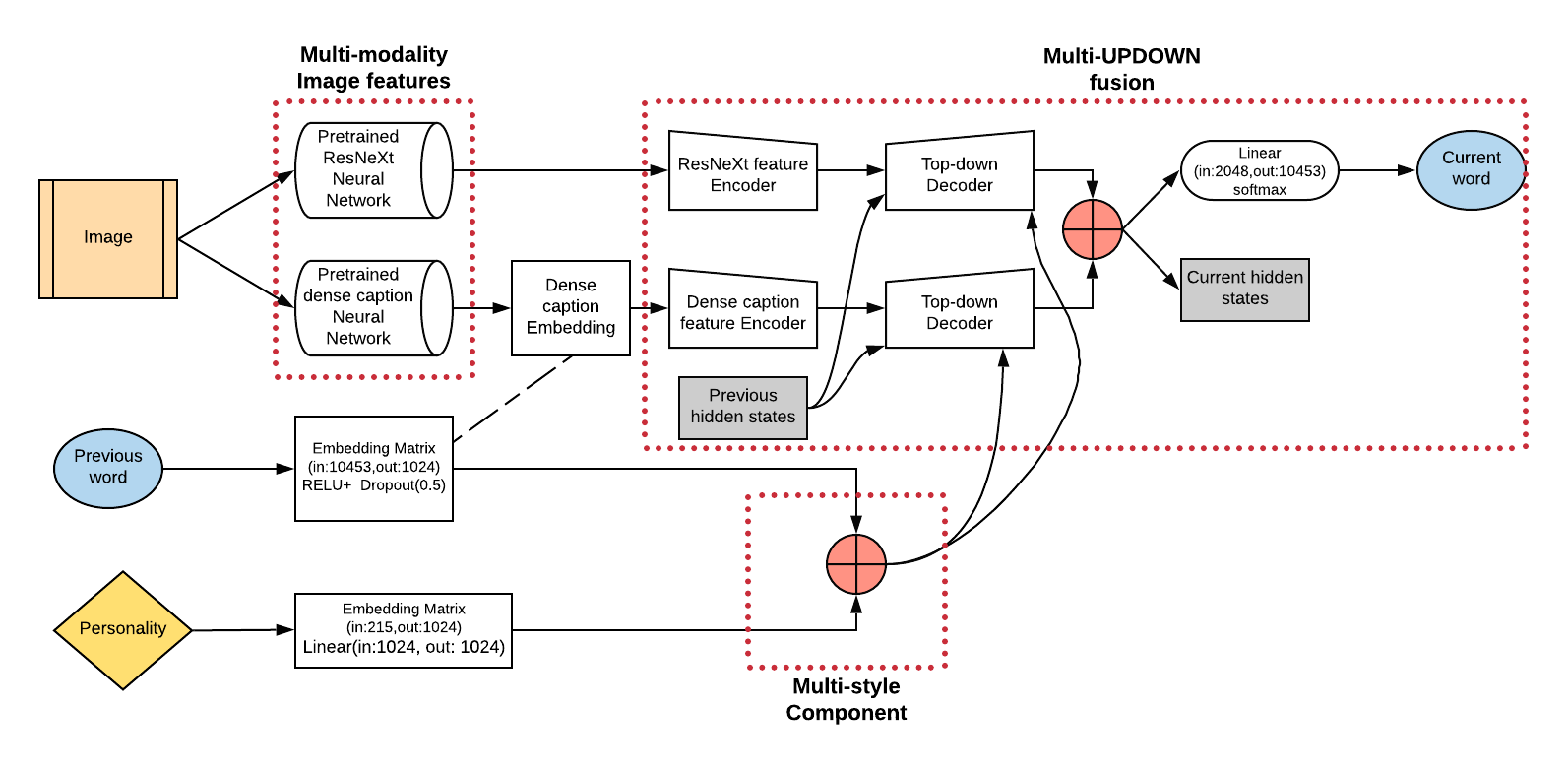}
\caption{Architecture for Multi-style image caption generation using Multi-modality features under Multi-UPDOWN model}
\label{fig:architecture}
\end{figure*}

\begin{figure*}[t]
\centering
\includegraphics[scale=0.45]{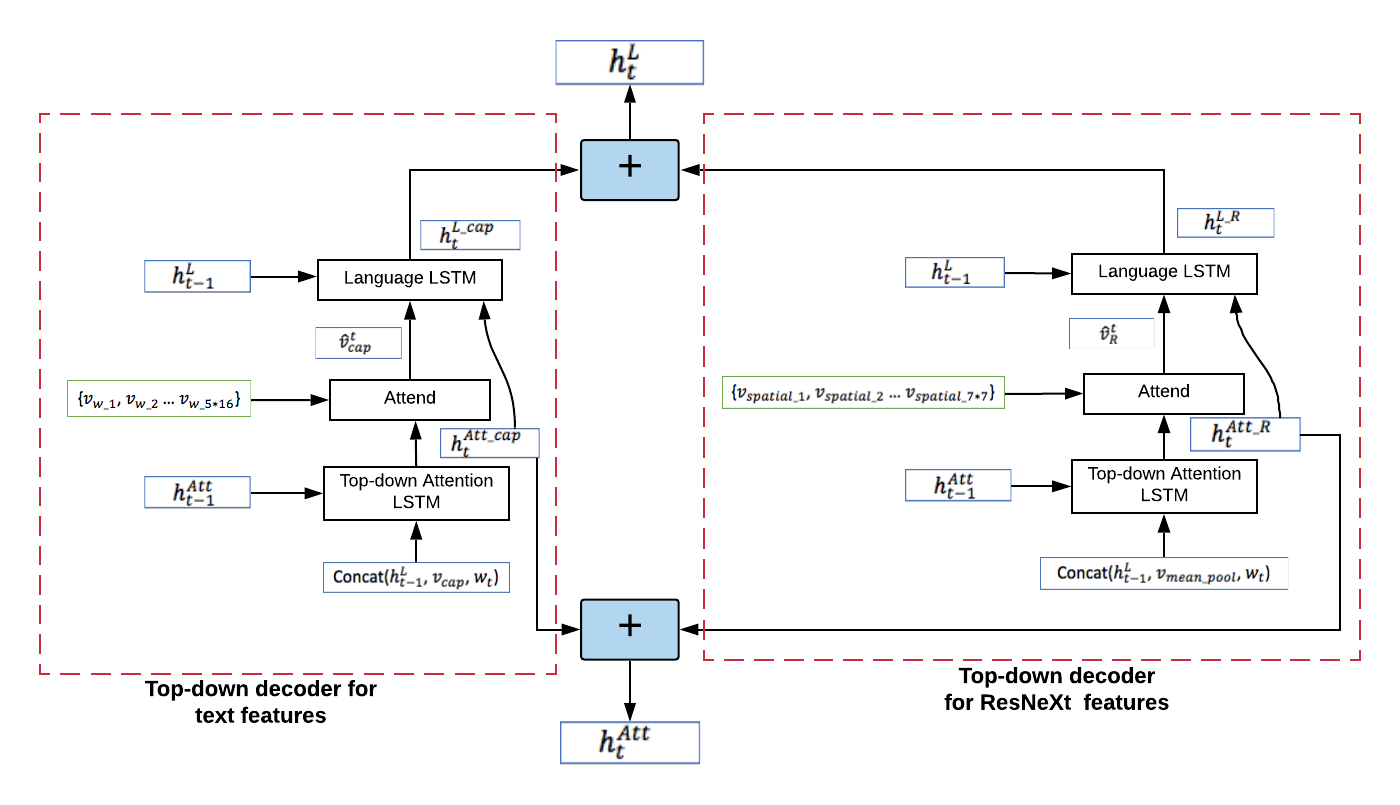}
\caption{Two Decoders Fusion Details}
\label{fig:two_updown}
\end{figure*}

\section{Related Work}

We first discuss work on stylish image captioning, and then on explanations for image captioning.

\subsection{Stylish Image Caption Model}
There has been a great deal of work on generating \textit{single style} captioning models~\cite{gan2017stylenet,chen2018factual,shuster2019engaging}. 
These models are designed to generate captions that exhibit a single style, such as Romantic or Humorous. 
As these models are limited to a single captioning style, they lack flexibility.

Later researchers explored developing models that addressed this limitation by enabling them to generate text in multiple possible styles \cite{guo2019mscap,zhao2020memcap} . 
Shuster \textit{et al.}~released the PERSONALITY-CAPTIONS dataset containing 215 personalities in 2019 for building engaging caption generations models. In their work, Shuster \textit{et al.}~built an image caption retrieval model and also explored the multi-style generative caption models along with various image encoding strategies
using several state-of-the-art image captioning models~\cite{xu2015show,anderson2018bottom}.
We extend the best performing supervised model presented in Shuster \textit{et al.}'s work, 
the \textbf{UPDOWN} model, to build a multi-style model which supports interpreting multi-modality image features. Due to the success of transformer structure on image captioning \cite{li2020oscar,zhang2021vinvl}, we also build a multi-style multi-modality image captioner by fine tuning on the pretrained model \cite{zhang2021vinvl}.

\subsection{Explanation for Image Caption Model}
There has been extensive work done on generating explanations for image captioning models. Many of these methods rely on identifying elements of the input or specific neurons in a neural net that significantly contribute to a generated caption \cite{sun2020understanding,amershi2015modeltracker,kang2018model}. 
In contrast, our model focuses on self-explanation and generating text explanations, which are easily human interpretable.
The goal of our work is to enable users with little-to-no experience in computer science or AI to understand the likely source of any errors in generated captions. 

\section{Methods}
We apply 2M ( multi-style multi-modality) on two popular deep learning structures to build stylish captioners. First, we extend the UPDOWN model to construct what we call a \textbf{3M} structure. Second, we build \textbf{2MT} by fine tuning a VinVL model and adjusting the input stream to generate stylish captions. We then outline how to make use of the 2M concept to explain the trained models when they generate faulty captions by using source error prediction.

\subsection{3M: Multi-style Multi-modality under Multi-UPDOWN Model}
The first contribution of this paper is an multi-fusion architecture that utilizes multi-modality fusion for performing multi-style image captioning. This architecture specifically utilizes the soft fusion of two parallel encoder-decoder blocks, with each block containing an UPDOWN-like attention module. 
Our overall architecture for one step generation can be seen in Figure~\ref{fig:architecture}, where our multi-UPDOWN fusion blocks synthesize the information from multi-modality image features, multi-style components (previous word, personality) and previous hidden states to predict current word and hidden states at each time step.

We utilize two features from pre-trained networks: ResNeXt \cite{xie2017aggregated} visual features and text features describing the image itself \cite{densecap}.
These features allow the learner to better ground the image features into natural language.
\subsubsection{Multi-style Component}
As shown in Figure \ref{fig:architecture}, the desired style of the output caption is given as an input to our system using a one-hot vector. 
We then use an embedding matrix $W_{p\_embed}$ and a linear layer to encode each style into a fixed-size vector, 
style vector $p$. 
For each word in our target stylized caption, we use another embedding matrix $W_{embed}$ to embed each word. We 
use $W_{embed}$ to embed the dense captions.
This enables us to better connect image features to natural language.
To better enable our network to generate words according to the given style, we concatenate each embedded word vector with the $p$ to create a stylized word vector, $\pmb{w}_{t}$.

\subsubsection{Multi-modality Image Features}
Our architecture relies on two sets of bottom-up features extracted using pre-trained networks: ResNeXt features and dense caption features. 
Specifically, we extract mean-pooled image features and spatial features from the ResNeXt network \cite{shuster2019engaging} and 5 dense captions from each image with a dense caption network \cite{densecap}. Each word in the dense captions is embedded using $W_{embed}$.
By collecting both visual and text features, we provide our architecture with a more complete understanding of the full context of the image. 

\subsubsection{Multi-UPDOWN fusion Model}
Our fusion model is composed of two individual encoders, the ResNext feature encoder and the dense caption encoder.
Our model also uses a fused Top-down fashion decoder, which used to decode captions from encoded image features.

\textbf{ResNeXt Feature Encoder and Dense Caption Encoder}
We encode the ResNeXt mean-pooled image features and spatial features using a linear layer, dropout layer and activation layer and get mean-pooled feature vector $\pmb{v}_{mean\_pool}$ and spatial feature vector $\pmb{v}_{spatial\_1}$, $\pmb{v}_{spatial\_2}$, 
..., $\pmb{v}_{spatial\_7*7}$. These are used as input features for the decoding process showed in the right branch of Figure \ref{fig:two_updown}. Then, we encode each embedded caption vector $Cap_{i}, i\in \{1,2,3,4,5\}$ using the Dense Caption Encoder, which is an LSTM network \cite{hochreiter1997long} shown below where $\pmb{w}^{dp}_{t,i}$ denotes a word vector in  $Cap_{i}$ at time t.
\begin{equation}\label{denseEncoder}
\footnotesize
\pmb{h}_{t,i}^{dp}, \pmb{c}_{t,i}^{dp}=LSTM(\pmb{w}^{dp}_{t,i},(\pmb{h}_{t-1,i}^{dp},\pmb{c}_{t-1,i}^{dp}))
\end{equation}
We concatenate all 5 hidden states $\pmb{h}_{i}^{dp}$ 
into one vector $\pmb{v}_{cap}$, which we call the \textit{caption vector}. 
To apply attention on specific words during the decoding procedure, we keep all word states $\pmb{c}_{t,i}^{dp}$  from the LSTM encoding process denoted as $\pmb{v}_{w_{1}},\pmb{v}_{w_{2}}...\pmb{v}_{w_{L}}$ where 5 captions contain total $L$ words.

\textbf{Top-down Decoder Fusion}
We apply the Top-down decoder model on encoded visual features and text features. 
At each time step, the Top-down decoder for text features generates a caption attention vector $\pmb{h}_{t}^{Att\_cap}$ by taking in the previous attention vector hidden states $\pmb{h}_{t-1}^{Att}$ as well as the concatenation of previous language model hidden states $\pmb{h}_{t-1}^{L}$, the caption vector $\pmb{v}_{cap}$ and the previous stylized word vector $\pmb{w}_{t}$ as input.
\begin{equation}
\label{TopDownAttLSTM}
\footnotesize
\pmb{h}_{t}^{Att\_cap}=TopDownAttLSTM([\pmb{h}_{t-1}^{L},\pmb{v}_{cap},\pmb{w}_{t}],\pmb{h}_{t-1}^{Att})
\end{equation}
To calculate the \textit{attended caption feature vector} We use a process inspired by \cite{anderson2018bottom}.
We use vectors $\pmb{v}_{w_{1}},\pmb{v}_{w_{2}}...\pmb{v}_{w_{L}}$ and the caption attention vector $\pmb{h}_{t}^{Att\_cap}$ in the below equations: 
\begin{align}\label{AttendAtt}
\footnotesize
& a_{i,t}= \pmb{w}_{a}^{T}tanh(W_{va}\pmb{v}_{w_{i}} + W_{ha}\pmb{h}_{t}^{Att\_cap}
) \\
& \pmb{\alpha}_{t} = softmax (\pmb{a}_{t}) \\
& \widehat{\pmb{v}}_{cap}^{t} =\sum_{i=1}^{K}{\pmb{\alpha}_{i}^{t}\pmb{v}_{w_{i}}}
\end{align}
where $W_{va} \in \mathbb{R}^{H\times V}, W_{ha}\in  \mathbb{R}^{H\times M}$ and $\pmb{w}_{a} \in \mathbb{R}^{H}$ are learned parameters.
This attention vector $\widehat{\pmb{v}}_{cap}^{t}$ is used as the input to the language LSTM layer where the initial state is the previous hidden state from the language model, $\pmb{h}_{t-1}^{L}$. 
This language LSTM then outputs the current language model hidden states $\pmb{h}_{t}^{L\_cap}$ for our text features as below:
\begin{equation}\label{LangLSTM}
\footnotesize
\pmb{h}_{t}^{L\_cap}=LanguageLSTM([\widehat{\pmb{v}}_{cap}^{t},\pmb{h}_{t}^{Att\_cap}],\pmb{h}_{t-1}^{L})
\end{equation}
We calculate the ResNeXt attention vector $\pmb{h}_{t}^{Att\_R}$, and current language model hidden states from ResNeXt features $\pmb{h}_{t}^{L\_R}$, using a similar process with a separate network (shown in Figure~\ref{fig:two_updown} right branch).
We generate the final language hidden states of the current step $\pmb{h}_{t}^{L}$ by fusing $\pmb{h}_{t}^{L\_cap}$, $\pmb{h}_{t}^{L\_R}$ as below:
\begin{equation}\label{eqfuse1}
\footnotesize
\pmb{h}_{t}^{L}= \pmb{h}_{t}^{L\_cap}+ \pmb{h}_{t}^{L\_R}
\end{equation}
We generate the final attention hidden states of the current step $\pmb{h}_{t}^{Att}$ by fusing  $\pmb{h}_{t}^{Att\_cap}$, $\pmb{h}_{t}^{Att\_R}$ as below:
\begin{equation}\label{eqfuse2}
\footnotesize
\pmb{h}_{t}^{Att}= \pmb{h}_{t}^{Att\_cap}+ \pmb{h}_{t}^{Att\_R}
\end{equation}
We get the final language output as below:
\begin{equation}\label{eqfuse3}
\footnotesize
\pmb{h}_{t}^{output}=Dropout(\pmb{h}_{t}^{L\_cap})+ Dropout(\pmb{h}_{t}^{L\_R})
\end{equation}
Then we apply a linear layer to project the final language output $\pmb{h}_{t}^{output}$ to the vocabulary space and use a log softmax layer to convert it to a log probability distribution.

\begin{figure*}[t]
\includegraphics[width=\textwidth,height=9cm]{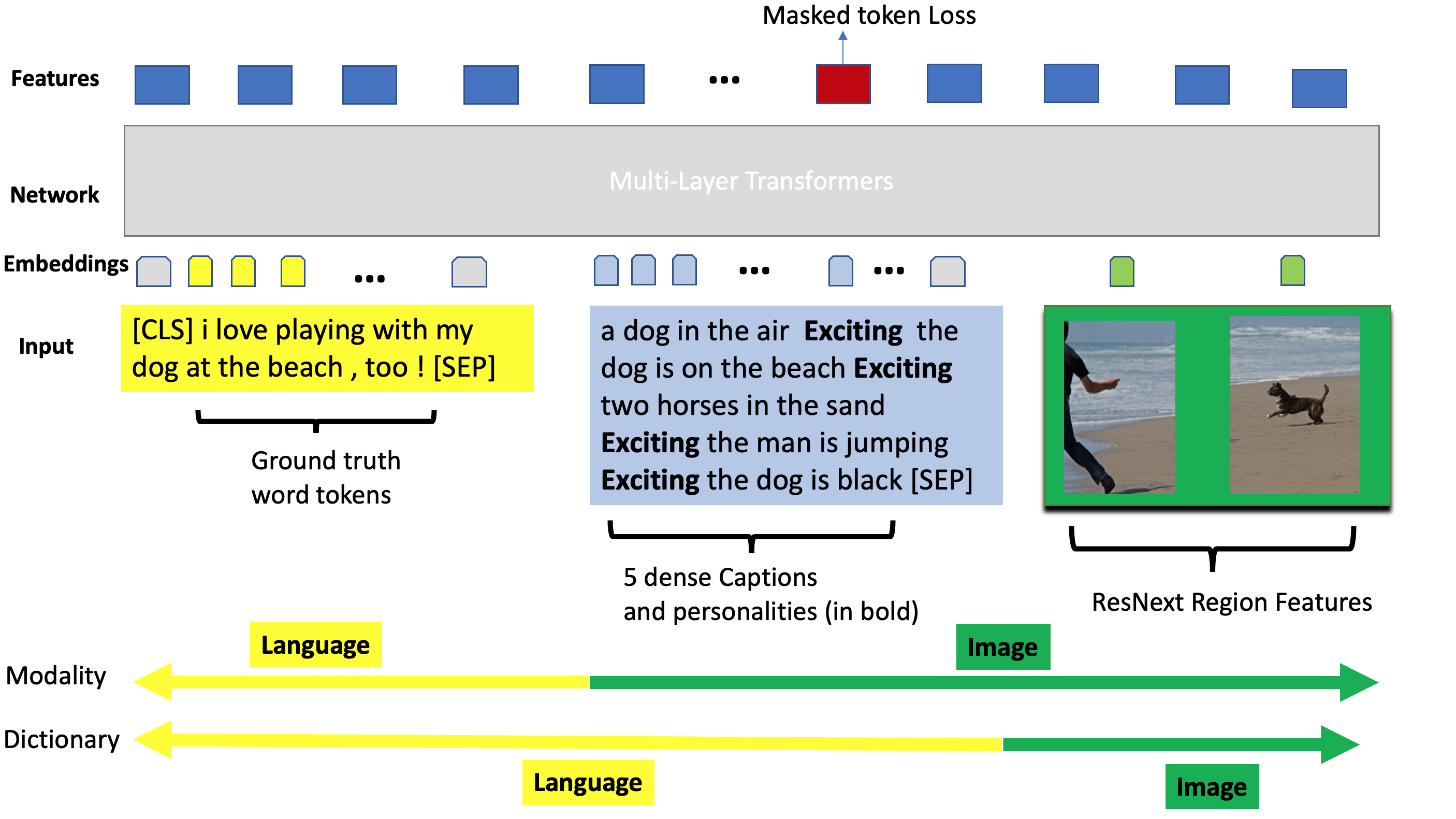}
\caption{The Framework of 2MT,showing the style modality with the fine-tuned on VinVL model. This figure is based on the figure from \cite{li2020oscar} extended to include our style components.}
\label{fig:vinvlstyle}
\end{figure*}
\subsection{2MT: Multi-style Multi-modality under Transformer Structure}


Recently, natural language transformers have emerged as an effective model for language generation. 
These massive networks are able to learn vast amounts of commonsense knowledge. 
Researchers typically harness this knowledge by \textit{fine tuning} these networks on more specialized data, hoping to combine general commonsense knowledge with more specialized domain knowledge.
Given this trend, we investigate if our 2M insights can be extended to transformer models for similar reasons. 

To investigate this, We built a multi-stylish image captioner, which we refer to as 2MT, out of the VinVL transformer model \cite{li2020oscar}. 
Our extended transformer architecture can be seen in Figure~\ref{fig:vinvlstyle}.
In VinVL, the input training data are pairs of $(w, q, v)$ where $w$ is the ground truth word tokens, $q$ is the set of detected object names (in text) and $v$ is the set of region features. The model aims to learn the relationship between captions and image region features by using the detected object names. 
As established earlier, however, factual image captioners that focus on region-based visual features may not be sufficient for performing stylish image captions. 
Thus, we need to extend this transformer model to take personality into consideration when generating text. 
Specifically, we exchange the anchor points $q$ in VinVL with the concatenation of a dense caption and personality indicator, $[densecap:personality]$, which are both in a text form. 
This is shown in the blue region of Figure~\ref{fig:vinvlstyle}.
In this way, the model learns relationships between text and visual features that are conditioned on personality. 
To further align our input, we also substitute $v$ with ResNext features (shown in the green region of Figure \ref{fig:vinvlstyle}). 
This ensures that the input information for both 3M and 2MT will be the same. 


We finetune on VinVL using masked token loss to learn the connection between image captions and image features, dense captions and personalities.  In this way, the language modality in Figure \ref{fig:vinvlstyle}, which includes dense captions, personality and the ground truth, share a common embedding space so that the model can learn which words are visually related and match the current style. In addition, ResNext and dense caption features come from the same image source, so their co-occurrence will help the model extract ResNext features associated with specific dense captions selected by style. 


\subsection{How does 2M Help Explain Multi-fusion Model?}
We have previously described two stylish captioners built using 2M. 
One of the reasons to use this technique is the expectation that it could provide diverse information that could enable error inference should the captioner produce erroneous text. Specifically, we use the multi-references provided by 2M, \textbf{(dense caption, current generation, other generations)+ground truth}, to infer erroneous input features. The ground truth is used to judge which part of the generated text is wrong. We will illustrate how to use 2M for explaining multi-fusion models through three steps: 1) How do we generate multi-references with 2M? 2) How can we align these multi-references? 3) How do we interpret the results of multi-reference alignment.  
\subsubsection{How do we Generate Multi-References with 2M?} 
For one image, with our multi-style trained model, we can generate text in many different styles by changing the input $p$. 
Each of these generated captions, along with dense captions and ground truth captions all describe the same images, but in different ways. 
By varying the input style, we can generate 4 sources of references (in text) for us to understand the relationship of the current generation with the inputs: 1) output caption from the current style $p$ (a sentence), 2) other captions generated using other styles (multiple sentences), 3) dense captions (5 sentences), and 4) the ground truth caption (5 sentences).

\subsubsection{How to Align the Multi-reference?}
\begin{figure*}
\centering
 \includegraphics[width=\textwidth,height=6cm]{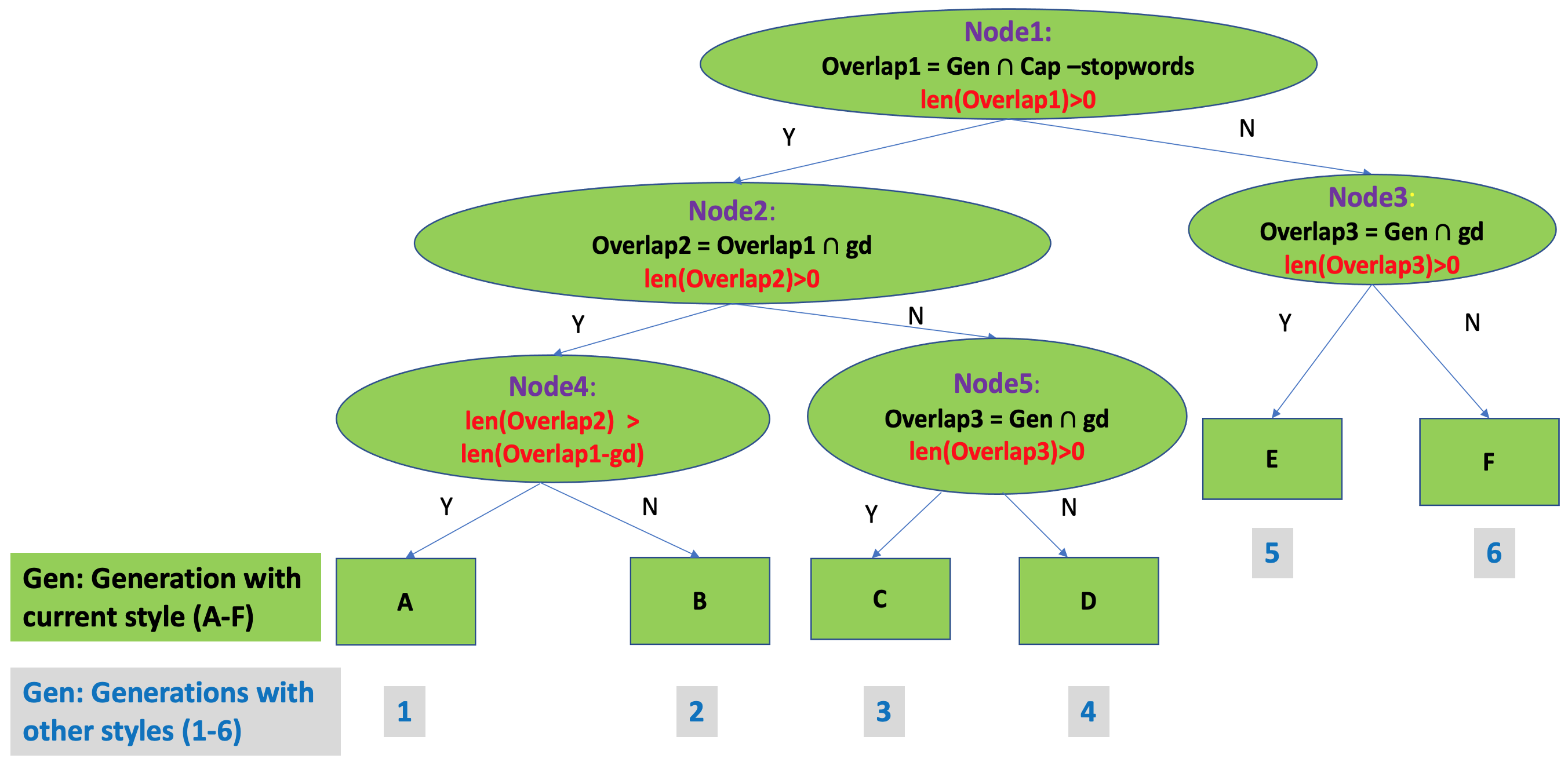}
\caption{Multi-View decision trees for estimating the potential feature errors. Gen:current generation or other generations words set; Cap: dense captions words set; stopwords:stop words from nltk library; gd: ground truth words set; $\cap$: intersections of two words sets.}
\label{fig:twotrees}
\end{figure*}

We use multi-view decision trees for aligning and comparing multi-references.
 The purpose of these trees is to mimic a heuristic that a humans could use to diagnose errors. 
 The trees are shown in Figure ~\ref{fig:twotrees}. First view results A-F were obtained by comparing a generated caption from the current style with the dense captions and the ground truth. Correspondingly, we get second view results 1-6 by comparing generated text  from other styles with the dense captions and the ground truth. The reasons we set the splitting point is as follows:
 
 \textbf{Node1} With this splitting point, we would like to see whether dense captions contribute to the generation or not. 
 
 \textbf{Node2} We would like to know whether the words from given visual information contributes to the performance or not.
 
 \textbf{Node3} We would like to explore whether ResNext features or style have positive contribution to the performance or not.
 
 \textbf{Node4} We would like to check whether the dense caption creates contributes noise to text generation. 
 
 \textbf{Node5} We would like to explore whether style or ResNext features positively contribute to the text generation if we already know some words from dense captions hinders performance.

\subsubsection{How do we Digest the Result from Multi-Reference Alignment?}
Three inputs (styles, Dense Captions, ResNext features) are the possible sources of errors when we perform error estimates. 
We combine results (A-F and 1-6) from the multi-view decision trees and create a check-table \ref{tab:rule} as estimation. Generally, we apply the following rules to estimate the error features in the table:\\
\textbf{Style is the error} When generated captions from other styles have better overlapping results with the ground truth or dense captions when compared to the captions generated using the current style (e.g., cells C-1, D-1, E-1, F-1 in Table \ref{tab:rule}); when visual features contribute to generated text (words from visual features are found in the generated text and in the ground truth), but performance of current generation is low, like cells A-1, A-2, A-3, A-4 in Table \ref{tab:rule};\\
\textbf{Dense captions are the error} When dense captions overlap with the current generation but those overlapping words are either not in ground truth, like cells C-4, D-2, D-4 in Table \ref{tab:rule} or there are fewer words in the ground truth than those that aren't in the ground truth, like cells B-2, B-4, C-2 in Table \ref{tab:rule};\\
\textbf{ResNext is the error} When the generated text has nonstop overlapping words with  the ground truth but these words are not in any of the dense captions, such like cells C-3, C-5, D-3, D-5 in Table \ref{tab:rule}. Using the two-view decision tree will help us to eliminate some bias where the overlapping words could come from current style, such as in cells E-1, F-1 where generations with other styles have good words overlapping with ground truth and dense captions, but generation with the current style does not. 
In this case, we will ascribe style is the error rather than ResNext or dense captions;\\
\textbf{Other is the error} We always give a second prediction as ``other'' when we found there is no other feature factor should be suspected as error source. Since we know it is possible other factors like model bias or dataset bias cause the error rather than features. But in this paper, we mainly focus on feature error predictions.


\begin{table*}[t]
 \centering
 \footnotesize
 \tabcolsep=0.10cm
 \begin{tabular}{lllllll}
  \hline
  {\bf OutputIndex}& {\bf 1}& {\bf 2} & {\bf 3}& {\bf 4} & {\bf 5}& {\bf 6}\\
  \hline
     A  & Style, Other & Style, Other & Style, Other &  ResNext, Other &  Style, Other&   ResNext, Other \\
  B & Style, Caption& Caption, Other & Style, Caption & Caption, Other & Style, Caption & ResNext, Caption\\
  C & Style, Caption & Caption, Other & ResNext, Other &  Caption, Other &  Caption, Style&   ResNext, Caption \\
  D  & Style, Caption & Caption, Other & Caption, Style &  Caption, Other &  Caption, Style&   ResNext, Caption\\
   E  & Style, Other & Caption, Other & Style, ResNext &  Caption, Other &  Style, other&  ResNext, Other\\
    F  & Style, Other & Style, Caption & Style, Caption &  Style, Caption & Style, ResNext&  Style, ResNext\\
 \hline
\end{tabular}
 \caption{Rule-based estimation based on multi-view decision tree outputs}
 \label{tab:rule}
\end{table*}
\section{Experiment Setting}
We performed experiments to verify the stylish captioning capabilities of our model as well as its usefulness in explaining model errors. 
\subsection{Multi-Style Captioning Model}
To demonstrate the effectiveness of our model on stylish image captioning, we use the PERSONALITY-CAPTIONS dataset, which contains 215 distinct personalities. We train them on 3M and 2MT, respectively.

We compare our results with the state-of-the-art work on the same datasets based on their automatic evaluation metrics. Ablation studies are also done on the 3M model to justify the contributions of each component of our method. 
To prove our model is expandable to linguistic stylized captions, we also train 3M on FlickrStyle10K dataset \cite{gan2017stylenet} which contains humorous and romantic personalities. We discuss the sample generations for two datasets in our qualitative studies.
\subsubsection{Dataset Details}

The ground truth captions in PERSONALITY-CAPTIONS \cite{shuster2019engaging,thomee2016yfcc100m} are created to be engaging and have a human-like style.
Each data entry in this dataset is represented as a triple containing an image, personality trait, and caption. 
In total, 241,858 captions are included in this dataset. 
In this work, we do not use the full PERSONALITY-CAPTIONS dataset due to accessibility of some examples. 
In total, our reduced dataset contains 186698 examples in the training set, 4993 examples in the validation set, and 9981 examples in the test set. The total vocabulary size of PERSONALITY-CAPTIONS after replacing infrequent tokens with 'UNK' is 10453. We perform replacement only when experimenting on 3M model. Since the tokenizer in Bert \cite{devlin2018bert} could directly mark infrequent tokens to unknown, so in the experiment of 2MT model, we directly use the original caption without any preprocessing.
The FlickrStyle10K dataset captions focus on linguistic style. Totally, 7000 images are publicly available. We trained 3M on FlickrStyle10K with the same splitting as \cite{guo2019mscap,zhao2020memcap}.


\subsubsection{Training and Inference}
In the training ofthe  3M model, we use entropy as a loss function and the Adam optimizer with an initial learning rate of 5e-4. 
The learning rate decays every 5 epochs. 
In total, we train 30 epochs with a batch size of 128 and evaluate the model every 3000 iterations. 
We train 30 epochs on 2MT model too. AdamW optimizer and linear scheduler are used and the initial learning rate is 3e-5.
We train for 100 epochs with batch size 128 when using the FlickrStyle10K dataset.

During inference, we generate captions using beam search with beam size 5. During this process, we impose a penalty to discourage the network of 3M from repeating words, from ending on words such as an, the, at, etc and from generating special tokens, like 'UNK'.

\subsubsection{Evaluation Methodology}
We perform both a quantitative and qualitative evaluation.
Our quantitative analysis is meant to show that our models can effectively generate stylish captions by outperforming state-of-the-art baselines on automated NLP metrics. 
In addition, we also run an ablation study on 3M model to validate the need for each part.\\
\textbf{Baselines and Evaluation Metrics}
We first evaluate the performance of the two models introduced here, the 3M model and the 2MT model. 
We compare them against the model introduced previously by Shuster \textit{et al.}~\cite{shuster2019engaging}. 
Since we use a subset of the original PERSONALITY-CAPTIONS dataset, we retrain the method outlined by Shuster \textit{et al.} using similar settings. 
We compare the performance of models using BLEU~\cite{papineni2002bleu}, ROUGE-L~\cite{lin-2004-rouge}, CIDEr~\cite{vedantam2015cider}, and SPICE~\cite{anderson2016spice}. The comparison results are listed in Table \ref{tab:PPerformance}.\\
\textbf{Ablation Study}
Additionally, to evaluate the benefits of each component of our model, we perform an ablation study using the PERSONALITY-CAPTIONS dataset. 
We compare the full 3M models against the following variations: no personality features, no text features, and no ResNeXt features. 
BLEU, ROUGE-L, CIDEr, and SPICE are reported in Table \ref{tab:Ablation} for evaluating the relevance between image and generations.
we also report the number of unique words used across all generated captions per model in Table \ref{tab:Ablation} to show the expressiveness of each generative model. 

Qualitatively, we seek to illustrate that our model is capable of generating captions that match the given style as well as the image context.
We first list the given image and five given dense captions, sample generations from 3M model along with personality in the parenthesis,  in Figure \ref{fig:rightwrong} as R1-R3.
We discuss the whether caption generations matching the context in three aspects: 1. whether the multi-style component working for connecting caption generations with given personality; 2. whether valid text features could help for generations to match the image; 3. whether ResNext feature could help make reasonable generations when the given text features fails to connect with the image. 
\begin{table*}[t]
\footnotesize
 \centering
 \tabcolsep=0.10cm
 \begin{tabular}{lllllllll}
 \hline
 \textbf {Method}& \textbf{Training Method}& \textbf{DenseCap} &\textbf{ResNeXt}& \textbf{ B1} & \textbf{ B4}& {ROUGE-L}& \textbf{CIDEr}& \textbf{SPICE}\\
 \hline
 UPDOWN \cite{shuster2019engaging}  & Supervised+REINFORCE & No & Yes & \bf 44.0 & 8.0 &27.4& 16.5 & \bf 5.2 \\
  UPDOWN \cite{shuster2019engaging}  & Supervised & No & Yes & 40.5 & 6.9 &26.2& 16.2 & 4.0 \\
 \bf 2MT & Supervised & Yes & Yes &  41.6 &   6.3&   26.8 &  15.2 & 4.8\\
  \bf 3M & Supervised & Yes & Yes &  43.0 & \bf 8.0&  \bf 27.6 & \bf 18.6 & 4.8\\
  \hline
\end{tabular}
 \caption{Performance of Generative Models on PERSONALITY-CAPTIONS Dataset. Note: Results of \cite{shuster2019engaging} under supervised learning are from re-training due to performance on supervised method not reported in \cite{shuster2019engaging} and some data of original dataset not available. We also listed original result of \cite{shuster2019engaging} which is under supervised and reinforcement learning for reference. B1-B4 denotes BLEU1-BLEU4.}
 \label{tab:PPerformance}
\end{table*}
\subsection{Examine the Explanation Capacity with Multi-reference from 2M}
We examine the explanation capacity of 2M on the test data of PERSONALITY-CAPTIONS. Specifically, we use multi-references to find the dominant erroneous features among multi-modality features so that we can use the found features to explain the errors.
We first define the examples which we think might have errors.
To do that, we see if the BLEU1 score of a test example is lower than the median BLEU1 score for the test data. 
These comprise the set of low performing examples, and we will attempt to identify the source of the errors present in these examples. 


When we generate multi-references, which involves regenerating captions under different styles, we choose the \textit{best} styles when making these replacements. 
Here, \textit{best} styles refer to those that are unlikely to have style errors. 
To determine this, we choose styles whose BLEU1 score is higher than the median BLEU1 score for the dataset. 
We utilize the top 5 best styles for both the 3M and 2MT model for this evaluation. When generating multi-references, we replace the current style with these 5 styles and use our models to generate new stylish captions. 
Using the original stylish caption generation, the 5 new stylish captions, the ground truth, and the dense captions used by each model, we estimate the likely source of the error using the multi-view decision tree described previously. 


We calculate the accuracy of the error estimate by comparing the predicted error result with the error feature ground truth. 
We list the calculation result for 3M model and 2MT model in the table \ref{tab:errorscore}. The ground truth error estimation is generated using error causal inference \cite{li2021error}.
We infer the feature error for each error example. The number of errors in the ground truth and the number of all the low performance examples are also reported in the table \ref{tab:errorscore}.

We also list the imperfect sample generations from 3M underlined in Figure \ref{fig:rightwrong} as W1-W2. With these imperfect generations and multiple generations under different styles, we will illustrate how we can use multi-reference for estimating input feature errors.
\section{Results and Discussion}
In this section, we will outline the results of our experiments and illustrate the caption model performance with respect to effectiveness and explainable capacity in quantitative and qualitative ways.

\subsection{Caption Model's Performance}

\textbf{Comparison with baselines and 2MT} As seen in Table \ref{tab:PPerformance}, our 3M under Multi-UPDOWN model outperforms single UPDOWN model under the same training method across all the NLP metrics we used for evaluation.We also achieve better results on ROUGE-L, CIDEr compared with Shuster's model trained under reinforcement learning. With different structure and same input information, 3M outperforms 2MT. 
Notably, 2MT didn't really gain benefits with multi-modality features compare to the single UPDOWN model, actually it has the similar performance as the single UPDOWN model. This might because we finetune on VinVL where the pre-training model is not pre-training with personality.
This provides evidence that 3M model is effective at multi-style caption generation. \\
\textbf{Ablation Study}
From Table \ref{tab:Ablation}, we can see if our model is trained without the multi-style component, the performance of all the nlp metrics drops, proving how critical this component is. 
Examining the results obtained from a model using only text features against a model that only had access to ResNeXt features shows that using only text features limits the overall expressiveness of generated captions as shown by the low number of unique words generated.

Our full model has achieved the highest ROUGE-L, CIDEr and SPICE score and improves expressiveness compared with model with only text features and improves the relevancy compared to a model with only Resnext features. \\
\textbf{Qualitative Analysis}
For our qualitative analysis, we will discuss the quality of the trained 3M models across two datasets assessing whether our model is capable of generating captions that match the given style and image context, and assessing whether our model can assist in finding reasons for imperfect captions. 

\begin{figure*}[!ht]
 \centering
 \includegraphics[width=\textwidth,height=8.5cm]{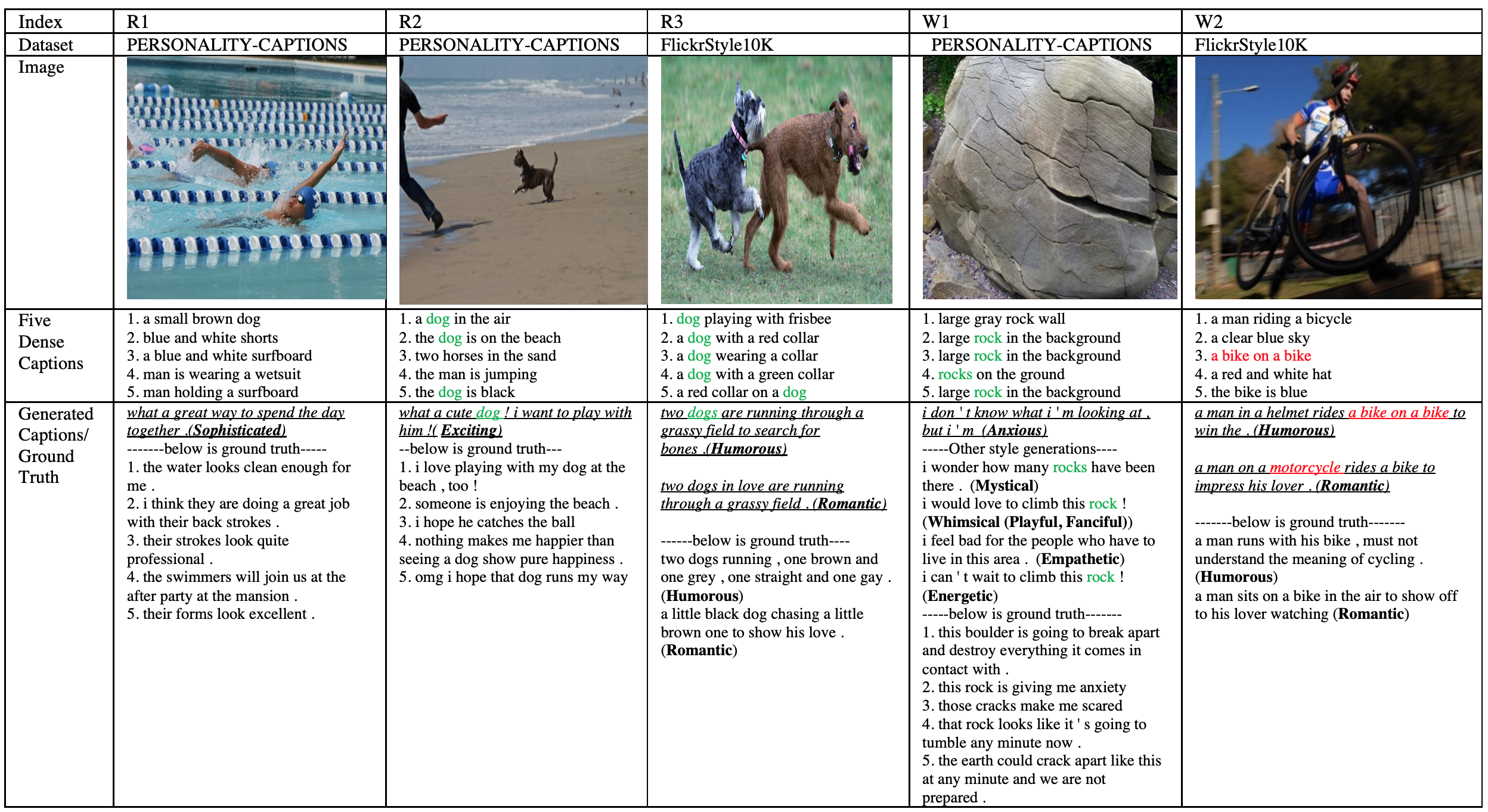}
 \caption{R1-R3: Generated Captions samples using 3M trained on PERSONALITY-CAPTIONS and FlickrStyle10K (underscored).
W1-W2: Imperfect Captions generations samples using 3M trained on PERSONALITY-CAPTIONS and FlickrStyle10K (underscored) along with generations from the same image and other personalities, personality are listed in parenthesis, ground truth has the same personality as the underscored generations}
 \label{fig:rightwrong}
\end{figure*}

From all generations in Figure \ref{fig:rightwrong}, we can see our 3M model is able to generate captions matching the given personality, which provides support that our multi-style component is able to help direct the generations in the desired personality tone. From R2-R3 we can see that when there is a valid text feature available, the 3M model could make use of them. The generation in R1 is expressed in a more conservative and global way since text features cannot provide correct information, which necessitates the use of ResNext features. 
\begin{table*}[t]
 \centering
 \footnotesize
 \tabcolsep=0.10cm
 \begin{tabular}{llllllllll}
  \hline
  {\bf Caption Model}& {\bf Personality}& {\bf DenseCap} & {\bf ResNeXt}& {\bf B1} & {\bf B4}& {\bf ROUGE-L}& {\bf CIDEr}& {\bf SPICE} &{\bf Unique words(\#)}\\
  \hline
     Multi-UPDOWN  & \bf No & Yes & Yes &  34.0 &  3.5&   22.3 &  11.1 & 3.6 & 257\\
  UPDOWN & Yes& \bf No & Yes & 42.4 & 7.5 &26.7& 17.9 & 4.4 & \bf 1558 \\
  UPDOWN & Yes & Yes & \bf No & \bf 43.2 & \bf 8.1&  \bf 27.6 &  18.0 &  4.6 & 1048\\
  Multi-UPDOWN  & Yes & Yes & Yes &  43.0 &  8.0&   \bf 27.6 & \bf 18.6 & \bf 4.8 & 1378 \\
 \hline
\end{tabular}
 \caption{Results of Ablation Studies on PERSONALITY-CAPTIONS Dataset}
 \label{tab:Ablation}
\end{table*}

\begin{table*}[t]
 \centering
 \footnotesize
 \tabcolsep=0.10cm
 \begin{tabular}{llll}
  \hline
  {\bf Model} &{\bf Accuracy} & {\bf Poor Performance Examples(\#)} & {\bf Single feature errors(\#)} \\
  \hline
     3M & 64.25\%  & 4982 & 1989 \\
  2MT& 84.26\% &  4988 & 4164 \\
 \hline
\end{tabular}
 \caption{Results of error estimation with multi-view tree; Single feature errors(\#) is the total error number of style, dense caption and ResNext features.}
 \label{tab:errorscore}
\end{table*}


\subsection{2M's Explanation Capacity} 
Here, we report the results of our quantitative and qualitative analysis on the capability of our models to aid in predicting errors. \\
\textbf{Quantitative Analysis}
From Table \ref{tab:errorscore}, we can see, assisted by 2M, the chance of human being able to estimate the dominant sourcing errors is more than 50\% on 3M model and 2MT model. 2M is especially good at assisting human finding errors on the 2MT model. This is likely because many of the erroneous examples generated by the 2MT model are single-feature errors. This is a stark contrast to the erroneous captions generated by 3M, which are mostly caused  by fusion feature errors. This tells us that the 2MT model has the capability of identifying the most important modality when generating captions, whereas the 3M model tends to fuse all modalities together when generating a caption. This makes it easier for humans to identify single feature errors from the 2MT model rather than the 3M model.\\
\textbf{Qualitative Analysis}
The example W1 in Figure \ref{fig:rightwrong} shows that extra references contain the right visual words "rock" while a caption generated using the style ``Anxious" is not even a complete sentence and we cannot see a complete view. With multi-references, we know the visual feature is correct while it is not interpreted correctly when combined with the style ``Anxious". This will not be clear if we only look at the generation with ``Anxious" and do single-reference inference, as we might ascribe the error to visual features. In W2, a human could easily recognize the bad phrase in the dense caption which also occurs in the generated caption with ``Humorous". If we change the style to ``Romantic", the visual attention will correct the visual word to ``bike" but add "motorcycle". This extra information showing visual information is not interpreted right under different styles, which confirm our thoughts that the visual information, especially dense captions, could be the error source for current generation.

\section{Conclusion}
In this paper we build two caption model: 3M and 2MT model supporting self-explaining, which are multi-style image captioner and could integrate multi-modal features and generate multiple stylish captions given one image.
We demonstrate the effectiveness of our 3M model by comparing against state-of-the-art work and 2MT model using automatic evaluation methods. Ablation studies have also be done to evaluate the contributions of each component of our 3M model. Since 3M and 2MT could provide multi-reference for an image, we also certify the multi-reference is helpful to explain the multi-modality fusion model on finding the dominant error features.


\bibliography{tacl2018}
\bibliographystyle{acl_natbib}

\end{document}